\pdfoutput=1

\documentclass[11pt]{article}

\usepackage[final]{acl}
\usepackage{times}
\usepackage{latexsym}
\usepackage{array}
\usepackage{afterpage}
\usepackage{colortbl}

\usepackage[T1]{fontenc}

\usepackage[utf8]{inputenc}

\usepackage{microtype}

\usepackage{inconsolata}

\usepackage{graphicx}
\usepackage{hyperref}

\title{CoVoSwitch: Machine Translation of Synthetic \\ Code-Switched Text Based on Intonation Units}

\author{Yeeun Kang \\
  Yale University \\
  \texttt{sophia.kang@yale.edu}}

\begin{document}
\maketitle
\begin{abstract}
Multilingual code-switching research is often hindered by the lack and linguistically biased status of available datasets. To expand language representation, we synthesize code-switching data by replacing intonation units detected through PSST, a speech segmentation model fine-tuned from OpenAI's Whisper, using a speech-to-text translation dataset, CoVoST 2. With our dataset, CoVoSwitch, spanning 13 languages, we evaluate the code-switching translation performance of two multilingual translation models, M2M-100 418M and NLLB-200 600M. We reveal that the inclusion of code-switching units results in higher translation performance than monolingual settings and that models are better at code-switching translation into English than non-English. Further, low-resource languages gain most from integration of code-switched units when translating into English but much less when translating into non-English. Translations into low-resource languages also perform worse than even raw code-switched inputs. We find that systems excel at copying English tokens but struggle with non-English tokens, that the off-target problem in monolingual settings is also relevant in code-switching settings, and that models hallucinate in code-switching translation by introducing words absent in both of the original source sentences. CoVoSwitch and code are available at \href{https://github.com/sophiayk20/covoswitch}{\url{https://github.com/sophiayk20/covoswitch}.}\footnote{ CoVoSwitch is released as a HuggingFace dataset. \href{https://huggingface.co/datasets/sophiayk20/covoswitch}{\url{https://huggingface.co/datasets/sophiayk20/covoswitch}}.}
\end{abstract}

\section{Introduction}

Code-switching (CSW), otherwise known as code-mixing, refers to the use of linguistic units from multiple languages in a conversation or utterance \citep{Pratapa:18}. In general, researching code-switching comprehensively is a complicated task due to the lack of code-switched data. One solution is to use existing code-switching datasets \citep{Weller:22, Nguyen:23}, but there is a limited number of such datasets and using them constrains research to the few language pairs that datasets are concentrated in, such as Spanish-English or Hindi-English \citep{Winata:23}. To alleviate the problem, previous work \citep{Alastruey:23} brought together multiple datasets, such as Fisher \citep{Cieri:04} and Bangor Miami \citep{Deuchar:14}. Nevertheless, in the multilingual setting, collecting data from multiple sources mixes different degrees of code-switching and blocks parallel understanding across languages.

Alternatively, most works have introduced synthetic datasets \citep{Winata:23}. These have been based on linguistic theories, such as the Matrix Language Frame (MLF) Model \citep{Myers-Scotton:97} and the Equivalence Constraint \citep{Poplack:80}. Applying the Equivalence Constraint requires the use of constituency parsers. \citep{Rizvi:21} utilized the Stanford Parser \citep{Klein:03} and the Berkeley Neural Parser \citep{Kitaev:18, Kitaev:19}. However, as of now, the Stanford Parser supports Arabic, Chinese, English, French, German, and Spanish, while the Berkeley Neural Parser supports Arabic, Basque, English, French, German, Hebrew, Hungarian, Korean, Polish, and Swedish. This presents a bottleneck in the number of languages that can be used for research and impedes the creation of code-switching data for unsupported or low-resource languages such as Tamil.

Synthetic datasets have also introduced code-switching mainly based on words. These include random replacements based on words \citep{Rijhwani:17, Xu:21, Rizvi:21, Tarunesh:21} and replacements based on connected components of aligned words \citep{Iyer:23}. However, word-based switching may not completely reflect the code-switching phenomenon. Recent research \citep{Pattichis:23} demonstrated that code-switching is more common across intonation units than within as a result of looser syntactic relationships and that intonation units should therefore serve as new replacement units instead of words. This constraint is referred to as the Intonation Unit Boundary Constraint.

To expand language representation, experiment with intonation units as basis units of code-switching, and reflect both linguistic and prosodic constraints, we synthesize data by following the Matrix Language Frame  Model and the Intonation Unit Boundary Constraint. We keep English as the matrix language and embed segments from non-English languages by replacing English intonation units of utterances from CoVoST 2 \citep{Wang:21}, a speech-to-text translation (S2TT) dataset, detected with PSST \citep{Roll:23}, an English prosodic speech segmentation model fine-tuned from OpenAI's speech recognition model Whisper \citep{Radford:23}. Utilizing S2TT datasets is advantageous for several reasons. First, they include transcripts for both languages and audio files for one language in each pair, which allows the simultaneous incorporation of text and speech features in code-switching data creation. Moreover, recent datasets cover a multitude of high-resource and low-resource languages, which enables the inclusion of diverse language pairs for synthetic code-switching data.

Meanwhile, we observe that while recent works \citep{Zhang:23, Khatri:23} have demonstrated the translation performance of multilingual large language models with billions of parameters such as XGLM-7.5B and BLOOMZ-7b1 on code-switching data, performance of multilingual neural machine translation (MNMT) models with millions of parameters remains relatively underexplored. We therefore measure the zero-shot code-switching translation performance of M2M-100 418M \citep{Fan:21} and NLLB-200 600M \citep{NLLBTeam:22}, capable of multilingual translation for 100 and 200 languages respectively, on our synthetic dataset.

Our contributions are summarized as follows: We (1) apply a single synthetic data generation method to different language pairs, including low-resource languages such as Tamil, based on a single dataset and thereby eliminate differences that emerge from the discrepancies in data generation methodology, (2) release a new code-switching dataset, CoVoSwitch, with similar code-switching levels across 13 languages, and (3) compare translation performance in code-switching versus monolingual settings and high-resource versus low-resource languages and identify the off-target problem and hallucinations. To the best of our knowledge, this is the first work to leverage prosodic segmentation features to create a dataset containing code-switched text.

\section{Synthetic Data Generation}

\begin{figure}
\centering
  \includegraphics[scale=0.25]{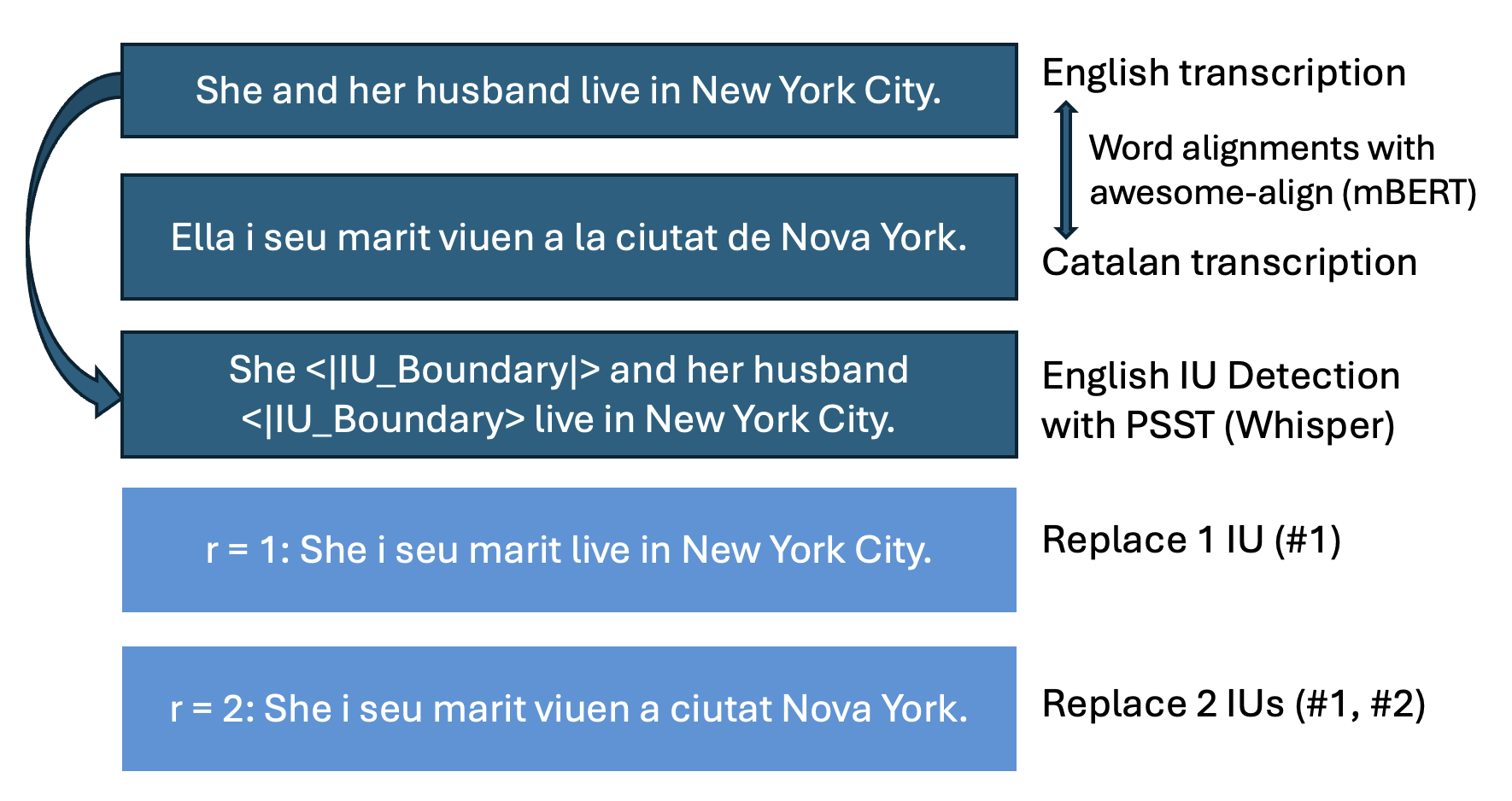}
  \caption{Our code-switching data generation pipeline with an example of English and Catalan parallel corpora.}
  \label{fig:datasynthesis}
\end{figure}

\renewcommand{\arraystretch}{1.2}

\begin{table}
    \small
    \centering
    \begin{tabular}{clll}
    \noalign{\global\arrayrulewidth=0.8pt}
    \hline
    \noalign{\global\arrayrulewidth=0.4pt}
    & Original & IU & Transcripts \\
    \hline
    Train & 289,413 & 195,166 & 100,176 \\
    \hline
    Valid. & 15,531 & 10,844 & 4,520 \\
    \hline
    Test & 15,531 & 9,252 & 3,688 \\
    \noalign{\global\arrayrulewidth=0.8pt}
    \hline
    \noalign{\global\arrayrulewidth=0.4pt}
    \end{tabular}
    \caption{Number of utterances used for dataset creation.}
    \label{tab:utterances_num}
\end{table}

\subsection{Intonation Unit Detection} 
We use the En$\rightarrow$X subset of the CoVoST 2 dataset, as this subset contains English recordings that we use to detect English prosodic boundaries. For non-English languages, we select Arabic (ar), Catalan (ca), Welsh (cy), German (de), Estonian (et), Persian (fa), Indonesian (id), Latvian (lv), Mongolian (mn), Slovenian (sl), Swedish (sv), Tamil (ta), and Turkish (tr). We follow the classification scheme of \citep{NLLBTeam:22} and denote Welsh, Mongolian, and Tamil as low-resource and others as high-resource. To match units of measurement for metrics such as CMI and SPF detailed later in this study, we exclude Chinese and Japanese, which are not whitespace separated. Further information on languages covered is contained in Appendix \ref{subsec:languages}. 

Using the PSST model\footnote{\href{https://github.com/nathan-roll1/psst}{\texttt{https://github.com/nathan-roll1/psst}}} \citep{Roll:23} fine-tuned from OpenAI's Whisper\footnote{\href{https://huggingface.co/openai/whisper-large-v3}{\texttt{https://huggingface.co/openai/whisper-large-v3}}} 
\citep{Radford:23}, we both generate transcriptions and detect intonation unit (IU) boundaries for English utterances in the original Common Voice 4.0 Corpus \citep{Ardila:20}, which serve as audio files for CoVoST 2. All English audio files were resampled at a sampling rate of 16,000 Hz to generate transcriptions with PSST. Of these, we extract sentences that contain intonation unit boundaries and exclude wrong transcriptions and outputs that contain hallucinations. Table \ref{tab:utterances_num} details the number of utterances used in each step, while Table \ref{tab:IUDetection} captures descriptive statistics on utterances used in the generated dataset.




\renewcommand{\arraystretch}{1.2} 

\begin{table}
    \centering
    \small
    \begin{tabular}{cccccc}
    \noalign{\global\arrayrulewidth=0.8pt}
    \hline
    \noalign{\global\arrayrulewidth=0.4pt}
    & & \textbf{$\mu$} & \textbf{$\sigma$} & min & max \\
    \hline
    \footnotesize{Train} & \footnotesize{IU} & 1.5 & 0.7 & 1 & 7 \\
        & words & 10.9 & 2.5 & 2 & 30 \\
    \hline
    Valid. & IU & 1.5 & 0.8 & 1 & 7 \\
        & words & 10.8 & 2.5 & 3 & 32 \\
    \hline
    Test & IU & 1.4 & 0.7 & 1 & 6 \\
        & words & 10.6 & 3.1 & 2 & 34 \\
    \noalign{\global\arrayrulewidth=0.8pt}
    \hline
    \noalign{\global\arrayrulewidth=0.4pt}
    \end{tabular}
    \caption{Statistics on English Common Voice intonation unit transcripts generated.}
    \label{tab:IUDetection}
\end{table}

\renewcommand{\arraystretch}{1} 

\subsection{Alignment Extraction and Intonation Unit Replacement}

We obtain word alignments between English and non-English text from CoVoST 2 using an aligner following previous research \citep{Rizvi:21, Winata:19, Pratapa:18}, but replace  \texttt{fast$\_$align} \citep{Dyer:13}, a reparametrization of IBM Model 2, with a neural aligner, \texttt{awesome-align}\footnote{\href{https://github.com/neulab/awesome-align}{\url{https://github.com/neulab/awesome-align}}} \citep{Dou:21}, because it outperforms \texttt{fast$\_$align} in alignment error rate. This aligner supports all target languages covered in this work as it is a fine-tuned aligner from mBERT \citep{Devlin:19}.

We pick the number of intonation units to replace, \textit{r}, from 1 to number of English intonation units - 1 for each English sentence. For each \textit{r}, we randomly select a combination of \textit{r} intonation unit indices, but nonconsecutive IU indices, if they exist, are prioritized over consecutive ones to represent more active code-switching. For each of the tokens in each replacement intonation unit selected, we find corresponding non-English tokens using word alignments. When replacing English tokens with non-English tokens, we preserve the original order in non-English languages. If no tokens are mapped by the aligner, empty strings are appended to the code-switched text, following previous work \citep{Pratapa:18}. For tokens that are not in the intonation units selected for replacement, English tokens are appended. Once the code-switched text is created, we perform checks to ensure that the synthesized text contains at least one intonation unit from both languages. Additionally, if the resulting code-switched text is exactly equal to the source English sentence, which occurs when tokens replaced are language-independent tokens such as proper nouns present in both component languages, we do not add the code-switched text to our dataset. Figure \ref{fig:datasynthesis} outlines an example synthesis process.

\subsection{Dataset Evaluation and Analysis}

\renewcommand{\arraystretch}{1.2} 

\begin{table}
    \centering
    \small
    \begin{tabular}{cccccc}
    \noalign{\global\arrayrulewidth=0.8pt}
    \hline
    \noalign{\global\arrayrulewidth=0.4pt}
         \textbf{ISO} & \textbf{Count} & \textbf{\%L1} & \textbf{\%L2} & \textbf{CMI} & \textbf{SPF} \\
        \hline
        ar & 5,176 & 55.20 & 44.80 & 32.89 & 0.17 \\
        ca & 5,137 & 51.02 & 48.98 & 33.54 & 0.16 \\
        cy & 5,150 & 52.37 & 47.63 & 33.32 & 0.16 \\
        de & 5,138 & 50.65 & 49.35 & 33.71 & 0.15 \\
        et & 5,153 & 55.71 & 44.29 & 32.76 & 0.17 \\
        fa & 5,174 & 52.07 & 47.93 & 33.43 & 0.16 \\
        id & 5,128 & 53.32 & 46.68 & 33.37 & 0.16 \\
        lv & 5,176 & 54.71 & 45.29 & 33.04 & 0.17 \\
        mn & 5,152 & 55.23 & 44.77 & 32.88 & 0.17 \\
        sl & 5,158 & 53.98 & 46.02 & 33.29 & 0.17 \\
        sv & 4,813 & 52.06 & 47.94 & 33.32 & 0.16 \\
        ta & 5,161 & 55.52 & 44.48 & 32.84 & 0.17 \\
        tr & 5,154 & 56.07 & 43.93 & 32.82 & 0.18 \\
    \noalign{\global\arrayrulewidth=0.8pt}
    \hline
    \noalign{\global\arrayrulewidth=0.4pt}
    \end{tabular}
    \caption{Test subset of CoVoSwitch. L1 is English, L2 is non-English language indicated by the ISO code.}
    \label{tab:syntheticdataset}
\end{table}

To evaluate our synthetic dataset, we report two automatic metrics, Code Mixing Index (CMI) and Switch Point Fraction (SPF). These metrics can be computed at either the utterance or corpus level, but we report at the corpus level to facilitate parallel understanding across languages.

CMI, first proposed by \citep{DasGamback:14}, measures the level of code-switching in a text. We follow the definition of \citep{Mondal:22} and report CMI as follows. For a code-switching sentence comprised of $\eta$ tokens, with $\eta_1$ and $\eta_2$ tokens in each language and $\eta = \eta_1 + \eta_2$, CMI is defined as 1 - $\frac{\max(\eta_1, \eta_2)}{\eta}$. We adhere to previous convention and multiply this number by 100. SPF was proposed by \citep{Pratapa:18} and measures the rate at which code-switching points occur in the code-switched text. SPF is defined as $\frac{\sum_{i=0}^{\eta-2}S(i, i+1)}{\eta-1}$ where $S(i, i+1)$ is an indicator variable that is equal to 1 if the tokens of indices $i$ and $i+1$ belong to different languages and else 0.

Table \ref{tab:syntheticdataset} captures information relevant to the test subset of our synthesized dataset, which is the only subset that we utilize in the experiments that follow. The total number of sentences generated is roughly 1.5 times the number of correct transcripts used in Table \ref{tab:utterances_num}, which is related to the average number of intonation units outlined in Table \ref{tab:IUDetection}. CMI values range from 32.76 to 33.71, which is comparable to CMI levels of 31.00 in \citep{Pratapa:18}. SPF values range from 0.15 to 0.18, which is comparable to SPF values of 0.17 and 0.2 in \citep{Winata:19}. Because our dataset is created by replacing entire intonation units instead of words as in previous works, it contains longer same language spans and less switch points, resulting in relatively higher CMI values and lower SPF values. In our dataset, roughly half of the tokens come from each constituent language. Statistics on train and validation subsets are included in Appendix \ref{subsec:trainvalidation}.

\section{Machine Translation Experimental Setup}

\begin{table*}[t]
    \centering
    \footnotesize
    \begin{tabular}{l|ccc|ccc|ccc}
    \noalign{\global\arrayrulewidth=0.8pt}
    \hline
    \noalign{\global\arrayrulewidth=0.4pt}
    & \multicolumn{3}{c|}{\textbf{spBLEU}} & \multicolumn{3}{c|}{\textbf{chrF++}} & \multicolumn{3}{c}{\textbf{COMET}} \\
    \hline
    \rowcolor{gray!20}
    \multicolumn{10}{c}{\textbf{X$\rightarrow$En}} \\
    \hline
    & \scriptsize{csw, En} & \scriptsize{M2M-100} & \scriptsize{NLLB-200} & \scriptsize{csw, En} &\scriptsize{M2M-100} & \scriptsize{NLLB-200} &
    \scriptsize{csw, En} & \scriptsize{M2M-100} & \scriptsize{NLLB-200} \\
    \hline
    ar & 38.2 & 31.8 & 41.1 
    & 48.4 & 55.5 & 61.3 
    & 72.1 & 81.1 & 85.2 \\
    
    ca & 43.6 & 41.5 & 50.2 & 
    \textbf{56.5} & 62.6 & 67.8
    & 75.2 & 83.1 & 86.7 \\
    
    cy & 41.8 & 9.4 & 46.8 
    & 54.5 & 30.0 & 65.2 
    & 67.2 & \underline{48.0} & 82.3 \\
    
    de & 40.0 & 38.0 & 47.5 
    & 55.8 & 60.3 & 66.3 
    & 77.4 & 83.9 & 88.1 \\
    
    et & 40.9 & 33.5 & 39.7 
    & 53.7 & 56.6 & 60.1 
    & 73.0 & 83.0 & 85.5 \\
    
    fa & 42.7 & 27.7 & 35.4 
    & 48.2 & 52.0 & 56.9 
    & 71.3 & 81.0 & 84.4 \\
    
    id & \textbf{46.1} & 36.2 & 46.2
    & 54.8 & 58.5 & 64.8 
    & \textbf{83.7} & 84.4 & 88.2 \\
    
    lv & 39.8 & 30.5 & 35.4 
    & 52.9 & 54.6 & 56.5 
    & 74.6 & 80.3 & 81.9 \\
    
    mn & 38.9 & \underline{9.1} & \underline{23.4} 
    & 47.9 & 30.5 & \underline{45.8} 
    & \underline{66.8} & 58.9 & \underline{77.5} \\
    
    sl & 42.4 & 34.1 & 42.7 
    & 53.8 & 57.2 & 62.6 
    & 74.7 & 82.2 & 86.3 \\
    
    sv & 43.5 & \textbf{44.5} & \textbf{51.9} 
    & 56.0 & \textbf{64.6} & \textbf{69.0} 
    & 83.0 & \textbf{85.6} & \textbf{88.9} \\
    
    ta & \underline{35.3} & \underline{9.1} & 38.2 
    & \underline{46.8} & \underline{29.8} & 59.4 
    & 71.1 & 59.1 & 86.1 \\
    
    tr & 41.3 & 28.3 & 37.0 
    & 52.9 & 52.3 & 57.9 
    & 71.7 & 82.4 & 86.2 \\
    
    \noalign{\global\arrayrulewidth=0.8pt}
    \hline
    \noalign{\global\arrayrulewidth=0.4pt}
    
    \rowcolor{gray!20}
    \multicolumn{10}{c}{\textbf{En$\rightarrow$X}} \\
    \hline
    & \scriptsize{csw, X} & \scriptsize{M2M-100} & \scriptsize{NLLB-200} & \scriptsize{csw, X} &\scriptsize{M2M-100} & \scriptsize{NLLB-200} &
    \scriptsize{csw, X} & \scriptsize{M2M-100} & \scriptsize{NLLB-200} \\
    \hline
    ar & 38.7 & 30.7 & 31.2 
    & 41.2 & 46.9 & 47.8 
    & 69.0 & 81.1 & 83.4 \\
    
    ca & 37.9 & 40.7 & 41.5 
    & 51.3 & 60.7 & 62.1 
    & 69.5 & 81.7 & 84.0 \\
    
    cy & 33.8 & \underline{2.3} & 29.8 
    & 45.1 & \underline{15.0} & 51.9 
    & 63.7 & \underline{36.8} & 78.5 \\
    
    de & 42.1 & 33.3 & 41.4 
    & \textbf{53.8} & 55.9 & 61.3 
    & 69.2 & 80.1 & 85.6 \\
    
    et & 42.5 & 28.7 & 27.0 
    & 51.7 & 51.9 & 50.9 
    & 70.2 & 82.8 & 83.0 \\
    
    fa & \underline{29.1} & 27.0 & 21.3 
    & \underline{36.4} & 44.7 & 39.2
    & 63.8 & 80.5 & 80.4 \\
    
    id & 39.2 & 36.6 & 43.2 
    & 51.9 & 61.0 & \textbf{65.6} 
    & \textbf{81.2} & \textbf{86.7} & \textbf{90.0} \\
    
    lv & 41.1 & 26.6 & 17.3 
    & 49.8 & 49.2 & 41.4 
    & 69.9 & 81.1 & \underline{72.9} \\
    
    mn & 32.3 & 2.9 & \underline{15.7} 
    & 38.5 & 17.6 & \underline{35.6}
    & \underline{61.5} & 50.8 & 79.2 \\
    
    sl & 39.5 & 32.2 & 32.4 
    & 49.7 & 53.1 & 53.7 
    & 68.8 & 82.7 & 84.4 \\
    
    sv & \textbf{42.7} & \textbf{44.4} & \textbf{46.5} 
    & \textbf{53.8} & \textbf{63.3} & 64.6 
    & 79.0 & 85.9 & 88.3 \\
    
    ta & 41.8 & 7.8 & 32.0 
    & 47.4 & 26.6 & 51.0 
    & 72.9 & 63.6 & 86.0 \\
    
    tr & 39.0 & 25.4 & 27.8 
    & 49.2 & 47.9 & 50.4 
    & 66.4 & 82.5 & 85.7 \\

    \noalign{\global\arrayrulewidth=0.8pt}
    \hline
    \noalign{\global\arrayrulewidth=0.4pt}
    
    \end{tabular}
    \caption{Metrics on raw code-switched inputs and monolingual translations, \textbf{best} and \underline{worst}.}
    \label{tab:baseline_monolingual}
\end{table*}

\begin{table*}[t]
    \centering
    \footnotesize
    \begin{tabular}{l|cccc|cccc|cccc}
    \noalign{\global\arrayrulewidth=0.8pt}
    \hline
    \noalign{\global\arrayrulewidth=0.4pt}
    
    & \multicolumn{4}{c|}{\textbf{spBLEU}} & \multicolumn{4}{c|}{\textbf{chrF++}} & \multicolumn{4}{c}{\textbf{COMET}} \\
    \hline
    \rowcolor{gray!20}
    & \multicolumn{2}{c}{\textbf{csw$\rightarrow$En}} & \multicolumn{2}{c|}{\textbf{csw$\rightarrow$X}} & 
    \multicolumn{2}{c}{\textbf{csw$\rightarrow$En}} & \multicolumn{2}{c|}{\textbf{csw$\rightarrow$X}} & 
    \multicolumn{2}{c}{\textbf{csw$\rightarrow$En}} & \multicolumn{2}{c}{\textbf{csw$\rightarrow$X}} \\
    \hline
    & \scriptsize{M2M} & \scriptsize{NLLB} & \scriptsize{M2M} & \scriptsize{NLLB} & 
    \scriptsize{M2M} & \scriptsize{NLLB} & \scriptsize{M2M} & \scriptsize{NLLB} &
    \scriptsize{M2M} & \scriptsize{NLLB} & \scriptsize{M2M} & \scriptsize{NLLB} \\
    \hline
    ar & 
    +22.7 & +24.8 & +7.0 & +14.0 & 
    +14.7 & +15.8 & +6.2 & +11.3 &
    +2.7 & +2.4 & +1.0 & +0.8 \\
    
    ca & 
    +4.5 & \underline{+18.9} & +13.7 & +7.5 & 
    -1.0 & +12.1 & +9.6 & +5.4 & 
    -9.5 & +1.6 & +0.4 & -2.9 \\
    
    cy &
    +27.0 & +19.6 & +12.8 & \underline{+2.7} & 
    +22.4 & \underline{+11.8} & +12.9 & \underline{+1.1} &
    \textbf{+12.8} & +1.8 & \textbf{+8.2} & -5.3 \\
    
    de & 
    \underline{+2.6} & +21.3 & +21.9 & +10.7 & 
    -1.3 & +13.6 & +14.5 & +7.1 & 
    -8.4 & +1.2 & +0.6 & -4.7
    \\
    
    et & 
    +4.1 & +24.0 & +21.9 & +11.9 & 
    -3.7 & +15.5 & +14.1 & +7.0 & 
    \underline{-13.8} & +0.9 & -0.8 & -5.1 
    \\
    
    fa & 
    +23.5 & +25.6 & \underline{+4.6} & +5.5 & 
    +15.2 & +16.0 & \underline{+3.7} & +4.1 & 
    +0.3 & +0.9 & -1.6 & -4.1
    \\
    
    id & 
    +12.0 & +22.8 & +19.3 & +14.7 & 
    +3.6 & +14.7 & +12.2 & +9.6 &
    -3.3 & +2.3 & +3.6 & +1.7
    \\
    
    lv & 
    +6.7 & +26.8 & \textbf{+25.0} & \textbf{+21.7} & 
    -1.4 & +18.0 & \textbf{+16.8} & \textbf{+15.1} &
    -9.3 & \textbf{+3.2} & +1.6 & \textbf{+3.5}
    \\
    
    mn & 
    +26.9 & \textbf{+28.2} & +12.4 & +7.1 &
    +21.1 & \textbf{+18.6} & +11.7 & +3.1 &
    +2.1 & +2.3 & +5.4 & -4.6
    \\
    
    sl & 
    +5.1 & +22.8 & +17.8 & +10.8 &
    \underline{-3.8} & +14.9 & +12.7 & +8.0 &
    -10.4 & +1.0 & -1.0 & -4.4
    \\
    
    sv & 
    +2.8 & +20.0 & +18.4 & +8.5 &
    -2.2& +13.0 & +12.2 & +6.1 &
    -5.4 & +1.9 & +1.8 & -2.6
    \\
    
    ta & 
    \textbf{+31.1} & +23.2 & +7.1 & +9.2 & 
    \textbf{+26.6} & +14.1 & +7.1 & +7.1 & 
    +11.2 & -0.1 & -1.1 & +0.2 \\
    
    tr & 
    +11.3 & +23.6 & +17.2 & +12.0 & 
    +2.1 & +15.0 & +11.4 & +8.0 & 
    -12.7 & \underline{-1.1} & \underline{-4.5} & \underline{-6.1} 
    \\
    
    \noalign{\global\arrayrulewidth=0.8pt}
    \hline
    \noalign{\global\arrayrulewidth=0.4pt}
    
    \end{tabular}
    \caption{Deltas of metrics on code-switching translations relative to monolingual translations in Table \ref{tab:baseline_monolingual}.}
    \label{tab:deltas_monolingual}
\end{table*}

\textbf{Models.} We use the HuggingFace pre-trained model checkpoints \texttt{facebook/m2m100\_418M} and \texttt{facebook/nllb-200-distilled-600M} for the M2M-100 418M and NLLB-200 600M models. These two models were chosen for their exceptional multilingual capabilities, with M2M-100 intended for non-English centric translation and NLLB-200 designed to improve translation performance in low-resource languages. Both support all languages covered by our synthetic dataset.

\noindent \textbf{Translation Settings.} We experiment with four translation settings for each of the English and non-English language pairs. First is csw$\rightarrow$En, in which code-switched text is translated into English. This setting was examined in previous research \citep{Nguyen:23, Xu:21}, but we also experiment with csw$\rightarrow$X to analyze any performance gaps that may arise by setting target language for translation differently. We compare these two code-switching translation settings to two monolingual translation settings, X$\rightarrow$En and En$\rightarrow$X, where X is a non-English language and En is English.

\noindent \textbf{Baselines.} Our baselines are twofold. First, we compare code-switching translations with monolingual translations and interpret deltas from monolingual baselines as the gains or losses from introducing code-switching units. We set our second baseline in consideration of our synthetic code-switched inputs. Because synthetic code-switched inputs already contain segments from reference texts, evaluation scores for these may be higher than translations of solely monolingual texts. In light of this, we consider deltas from raw code-switched inputs the performance of systems in translating code-switched text.

\noindent \textbf{Evaluation Metrics.} We measure the performance of translation models with the following automatic metrics: chrF++ \citep{Popovic:17} at the character level, spBLEU \cite{Goyal:22} at the language-agnostic subword level tokenized through SentencePiece \citep{Kudo:18}, and COMET \citep{Rei:20} at the detokenized representation level. spBLEU and chrF++ measure similarity between reference translation and system translation, while COMET predicts human judgments of system translations based on a neural model. We use the \texttt{FLORES-200} \citep{NLLBTeam:22} tokenizer available through SacreBLEU \citep{Post:18} for spBLEU and \texttt{Unbabel/wmt22-comet-da} \citep{Rei:22} for COMET calculation.

We supplement chrF++, spBLEU, and COMET with copy and replacement rates to examine whether translation systems can perform implicit language identification to copy or replace tokens as appropriate. As in \citep{Liu:21, Xu:21, Song:19}, we define copy rate as the rate at which the target tokens already present in code-switched input is successfully transferred over to the machine translation system output. We define replacement rate as the rate at which the system successfully converts non-target input tokens to target tokens. It follows that lower replacement rates indicate less translated outputs.

All experiments are conducted on a single NVIDIA L4 GPU. 

\section{Results and Discussion}

\subsection{Code-Switched Inputs Relative to Monolingual Translations} 

\begin{table*}
    \centering
    \footnotesize
    \begin{tabular}{l|cccc|cccc|cccc}

    \noalign{\global\arrayrulewidth=0.8pt}
    \hline
    \noalign{\global\arrayrulewidth=0.4pt}
    
    & \multicolumn{4}{c|}{\textbf{spBLEU}} & \multicolumn{4}{c|}{\textbf{chrF++}} & \multicolumn{4}{c}{\textbf{COMET}} \\
    \hline
    \rowcolor{gray!20}
    & \multicolumn{2}{c}{\textbf{csw$\rightarrow$En}} & \multicolumn{2}{c|}{\textbf{csw$\rightarrow$X}} & 
    \multicolumn{2}{c}{\textbf{csw$\rightarrow$En}} & \multicolumn{2}{c|}{\textbf{csw$\rightarrow$X}} & 
    \multicolumn{2}{c}{\textbf{csw$\rightarrow$En}} & \multicolumn{2}{c}{\textbf{csw$\rightarrow$X}} \\
    \hline
    & \scriptsize{M2M} & \scriptsize{NLLB} & \scriptsize{M2M} & \scriptsize{NLLB} & 
    \scriptsize{M2M} & \scriptsize{NLLB} & \scriptsize{M2M} & \scriptsize{NLLB} &
    \scriptsize{M2M} & \scriptsize{NLLB} & \scriptsize{M2M} & \scriptsize{NLLB} \\
    \hline
    ar & 
    \textbf{+16.3} & +27.7 & -1.0  & +6.5 &
    \textbf{+21.8} & \textbf{+28.7} & +11.9 & +17.9 &
    \textbf{+11.7} & +15.5 & +13.1 & \textbf{+15.2} \\
    
    ca & 
    +2.4 & +25.5 & +16.5 & +11.1 & 
    +5.1 & +23.4 & +19.0 & +16.2 &
    -1.6 & +13.1 & +12.6 & +11.6 \\
    
    cy &
    \underline{-5.4} & +24.6 & -18.7 & -1.3 &
    \underline{-2.1} & +22.5 & \underline{-17.2} & +7.9 &
    \underline{-6.4} & +\textbf{16.9} & \underline{-18.7} & +9.5 \\
    
    de & 
    +0.6 & \textbf{+28.8} & +13.1 & +10.0 &
    +3.2 & +24.1 & +16.6 & +14.6 &
    -1.9 & +11.9 & +11.5 & +11.7 \\
    
    et & 
    -3.3 & +22.8 & +8.1 & -3.6 &
    -0.8 & +21.9 & +14.3 & +6.2 &
    -3.8 & +13.4 & +11.8 & +7.7 \\
    
    fa & 
    +8.5 & +18.3 & +2.5 & -2.3 &
    +19.0 & +24.7 & +12.0 & +6.9 &
    +10.0 & +14.0 & \textbf{+15.1} & +12.5 \\
    
    id & 
    +2.1 & +22.9 & +16.7 & \textbf{+18.7} &
    +7.3 & +24.7 & +21.3 & \textbf{+23.3} &
    -2.6 & \underline{+6.8} & +9.1 & +10.5 \\
    
    lv & 
    -2.6 & +22.4 & +10.5 & -2.1 & 
    +0.3 & +21.6 & +16.2 & +6.7 &
    -3.6 & +10.5 & +12.8 & \underline{+6.5} \\
    
    mn & 
    -2.9 & \underline{+12.7} & -17.0 & \underline{-9.5} &
    +3.7 & \underline{+16.5} & -9.2 & \underline{+0.2} &
    -5.8 & +13.0 & -5.3 & +13.1 \\
    
    sl & 
    -3.2 & +23.1 & +10.5 & +3.7 &
    -0.4 & +23.7 & +16.1 & +12.0 &
    -2.9 & +12.6 & +12.9 & +11.2 \\
    
    sv & 
    +3.8 & +28.4 & \textbf{+20.1} & +12.3 &
    +6.4 &  +26.0 & \textbf{+21.7} & +16.9 & 
    -2.8 & +7.8 & +8.7 & +6.7 \\
    
    ta & 
    +4.9 & +26.1 & \underline{-26.9} & -0.6 &
    +9.6 & +26.7 & -13.7 & +10.7 &
    -0.8 & +14.9 & -10.4 & +13.3 \\
    
    tr & 
    -1.7 & +19.3 & +3.6 & +0.8 &
    +1.5 & +20.0 & +10.1 & +9.2 &
    -2.0 & +13.4 & +11.6 & +13.2 \\

    \noalign{\global\arrayrulewidth=0.8pt}
    \hline
    \noalign{\global\arrayrulewidth=0.4pt}
    
    \end{tabular}
    \caption{Deltas of metrics on code-switching translations relative to raw code-switched inputs in Table \ref{tab:baseline_monolingual}.}
    \label{tab:deltas}
\end{table*}

Results are shown in Table \ref{tab:baseline_monolingual}. Inspection of spBLEU in the to English setting reveals that 12 out of 13 synthetic code-switched inputs score higher than M2M-100 translation outputs when evaluated against reference English texts. For NLLB-200, however, only 5 code-switched inputs score higher than monolingual translations. In contrast, in the to non-English setting, raw inputs score higher than monolingual translations for 11 and 10 languages. We thus reaffirm the findings of \citep{Nguyen:23} that code-switched inputs score higher than monolingual translations but with qualifications that exceptional monolingual translations by stronger models can outperform code-switched inputs and that this assertion holds more true for the to non-English setting than the to English setting.

Further, we observe that in spBLEU and chrF++ for low-resource languages such as Welsh, Mongolian, and Tamil, gaps between scores for raw code-switched inputs and monolingual translations are larger, mainly due to worse performance of models in translating these languages. M2M-100 struggles with translation across all three languages, while NLLB-200 shows better translations. COMET scores similarly suggest that M2M-100 shows weak performance in Welsh, Mongolian, and Tamil, as they are the only languages with COMET scores under 80 in both monolingual translation settings.

\subsection{Deltas Relative to Monolingual Baselines} 
\textbf{Inclusion of code-switched units results in better translation than monolingual settings.} This is seen in the predominantly positive deltas across spBLEU and chrF++ in Table \ref{tab:deltas_monolingual}. In particular, whether the languages are low-resource or high-resource, spBLEU scores increase across all languages, models, and translation settings. We notice similar trends in chrF++ with all scores increasing for csw$\rightarrow$X. For csw$\rightarrow$En, some minimal decreases are observed for M2M-100 in high-resource languages, while all scores increase for NLLB-200. However, improvements can be made, as deltas for COMET scores are smaller than in other metrics.

\noindent \textbf{Low-resource languages gain most in csw$\rightarrow$En and but much less in csw$\rightarrow$X.} In csw$\rightarrow$En translation in Table \ref{tab:deltas_monolingual}, low-resource languages benefit the most with two-digit gains from monolingual translations, whereas high-resource languages show smaller gains. This is most prominent in M2M-100 when translating into English. Tamil, Welsh, and Mongolian show the most gains with spBLEU increases of 31.1, 27.0, and 26.9 each, while German and Swedish increase by 2.6 and 2.8. Welsh for NLLB-200 is ranked penultimately, but we regard this as trivial as spBLEU scores for NLLB-200 have a very high average gain of 23.2 and a low standard deviation of 2.7. However, for low-resource csw$\rightarrow$X translation, gains from monolingual are much smaller than in csw$\rightarrow$En. In M2M-100, csw$\rightarrow$X deltas are halved or more than halved from csw$\rightarrow$En deltas for Welsh, Mongolian, and Tamil, while csw$\rightarrow$X deltas become significantly larger for high-resource languages such as German, Estonian, and Latvian. In NLLB-200 csw$\rightarrow$X translation, all low-resource languages show one digit spBLEU and chrF++ deltas. NLLB-200 benefits particularly little in Welsh given the 2.7 increase in spBLEU and 1.1 increase in chrF++. This extends findings of \citep{Goyal:22} that translating into low-resource languages is harder than translating out of them. Table \ref{tab:topgains} summarizes two languages with the most and least gains in spBLEU for each model and setting.

\begin{table}[h]
    \small
    \begin{tabular}{lcccc}
    
    \noalign{\global\arrayrulewidth=0.8pt}
    \hline
    \noalign{\global\arrayrulewidth=0.4pt}

        \rowcolor{gray!20}
        & \multicolumn{2}{c}{\textbf{csw$\rightarrow$En}} & \multicolumn{2}{c}{\textbf{csw$\rightarrow$X}} \\
        \hline
        & \scriptsize{M2M-100} & \scriptsize{NLLB-200} & \scriptsize{M2M-100} & \scriptsize{NLLB-200} \\
        \hline
        \footnotesize{$\uparrow$} & ta (+31.1) &  mn (+28.2) & lv (+25.0) & lv (+21.7) \\
            & cy (+27.0) & lv (+26.8) & de (+21.9) & id (+14.7) \\
        \hline
        & sv (+2.8) & cy (+19.6) & ar (+7.0) & fa (+5.5) \\
        \footnotesize{$\downarrow$} & de (+2.6) &  ca (+18.9) & fa (+4.6) & cy (+2.7) \\

    \noalign{\global\arrayrulewidth=0.8pt}
    \hline
    \noalign{\global\arrayrulewidth=0.4pt}
    
    \end{tabular}
    \caption{Languages with most and least spBLEU gain by introduction of code-switching relative to monolingual.}
    \label{tab:topgains}
\end{table}

\subsection{Deltas Relative to Code-Switched Input Baselines} 
\textbf{Models are better in code-switching translation into English than non-English.} \citep{Goyal:22} established that multilingual translation models are better at translation into English than into non-English languages. We confirm similar results in code-switching settings. This is most evident in Table \ref{tab:deltas} with gains in performance for chrF++ and spBLEU for NLLB-200, where differences in deltas between csw$\rightarrow$En and csw$\rightarrow$X are double digits for the majority of the languages. 

\noindent \textbf{High-resource languages gain further while low-resource languages lose performance gained through code-switched inputs in csw$\rightarrow$X.}
Performance already gained from code-switched input is lost in low-resource languages for csw$\rightarrow$X translation, whereas translations for high-resource languages effectively use code-switched inputs to result in even greater gains than those seen in csw$\rightarrow$En translation. For instance, deltas of chrF++ scores in M2M-100 Catalan translation are 5.1 in csw$\rightarrow$En and 19.0 in csw$\rightarrow$X, compared to values in Welsh of -2.1 in csw$\rightarrow$En and -17.2 in csw$\rightarrow$X. Similar sized drops are seen for csw$\rightarrow$X in Tamil with -13.7 and Mongolian with -9.2. Comparatively, NLLB-200 performs better, but the increase in csw$\rightarrow$X in Mongolian is a mere 0.2 compared to 23.3 in Indonesian. NLLB-200 spBLEU scores yield similar conclusions, with a drop of 9.5 observed in Mongolian compared to an increase of 18.7 in Indonesian and 12.3 in Swedish. Overall, negative deltas for csw$\rightarrow$X translation suggest that there is room for improvement for code-switching translation into non-English languages.

\subsection{Analysis of Translations}

\begin{table}
    \centering
    \small
    \begin{tabular}{lcccc}

    \noalign{\global\arrayrulewidth=0.8pt}
    \hline
    \noalign{\global\arrayrulewidth=0.4pt}
    
    \hline
    \rowcolor{gray!20}
    & \multicolumn{2}{c}{\textbf{csw$\rightarrow$En}} & \multicolumn{2}{c}{\textbf{csw$\rightarrow$X}} \\
    \hline
    & \scriptsize{M2M-100} & \scriptsize{NLLB-200} & \scriptsize{M2M-100} & \scriptsize{NLLB-200} \\
    \hline
    ar & 92.8 & 97.9 & 69.0 & 80.8 \\
    ca & 94.2 & 96.3 & 92.3 & 83.1 \\
    cy & 93.9 & 98.4 & 54.0 & 70.2 \\
    de & 94.1 & 96.2 & 93.3 & 83.1 \\
    et & 94.2 & 96.2 & 88.6 & 68.5 \\
    fa & 93.0 & 96.8 & 70.2 & 65.6 \\
    id & 94.1 & 97.5 & 94.2 & 92.9 \\
    lv & 93.9 & 96.8 & 93.0 & 77.0 \\ 
    mn & 90.4 & 95.5 & 51.0 & 55.5 \\
    sl & 94.0 & 96.8 & 89.7 & 75.9 \\
    sv & 94.5 & 97.0 & 95.4 & 83.0 \\ 
    ta & 91.0 & 96.7 & 37.7 & 69.4 \\
    tr & 94.0 & 96.5 & 83.3 & 76.3 \\

    \noalign{\global\arrayrulewidth=0.8pt}
    \hline
    \noalign{\global\arrayrulewidth=0.4pt}
    
    \end{tabular}
    \caption{Copy rates (\%) of code-switching translations.}
    \label{tab:copyrate}
\end{table}

\textbf{Copy Rates.}
We report copy rates in Table \ref{tab:copyrate}. For csw$\rightarrow$En translation, models show high copy rates ranging from 90.4 to 94.5 percent for M2M-100 and 95.5 to 98.4 percent for NLLB-200. This is in line with findings of \citep{Xu:21} in which high copy rates were observed for csw$\rightarrow$En translations, with code-switched text created using English, French, and Spanish. Conversely, for csw$\rightarrow$X, models show less competent copy rates. In particular, M2M-100 exhibits copy rates of around only 50 percent for Welsh and Mongolian and below 50 percent for Tamil. NLLB-200 obtains better performance with Welsh and Tamil, but still shows weak performance for Mongolian at 55.5 percent. Copy rates for csw$\rightarrow$X are worse than csw$\rightarrow$En for every language and model except for M2M-100 in Indonesian and Swedish. 

\begin{table}[t]
    \centering
    \small
    \begin{tabular}{lcccc}

    \noalign{\global\arrayrulewidth=0.8pt}
    \hline
    \noalign{\global\arrayrulewidth=0.4pt}
    
    \hline
    \rowcolor{gray!20}
    & \multicolumn{2}{c}{\textbf{csw$\rightarrow$En}} & \multicolumn{2}{c}{\textbf{csw$\rightarrow$X}} \\
    \hline
    & \scriptsize{M2M-100} & \scriptsize{NLLB-200} & \scriptsize{M2M-100} & \scriptsize{NLLB-200} \\
    \hline
    ar & 100.0 \scriptsize{(0.0)} & 100.0 \scriptsize{(0.0)} & 99.9 \scriptsize{(0.0)} & 100.0 \scriptsize{(0.0)} \\
    
    ca & 75.7 \scriptsize{(-6.4)} & 96.3 \scriptsize{(-2.2)} & 92.3 \scriptsize{(-3.3)} & 83.1 \scriptsize{(-3.2)} \\
    
    cy & 69.8 \scriptsize{(-6.7)} & 77.9 \scriptsize{(-0.7)} & 87.2 \scriptsize{(-1.3)} & 89.2 \scriptsize{(-2.1)} \\

    de & 100.0 \scriptsize{(0.0)} & 100.0 \scriptsize{(0.0)} & 89.8 \scriptsize{(-2.0)} & 89.8 \scriptsize{(-2.0)} \\

    et & 100.0 \scriptsize{(0.0)} & 100.0 \scriptsize{(0.0)} & 82.1 \scriptsize{(-3.5)} & 82.2 \scriptsize{(-4.0)} \\

    fa & 100.0 \scriptsize{(0.0)} & 100.0 \scriptsize{(0.0)} & 99.9 \scriptsize{(0.0)} & 100.0 \scriptsize{(0.0)} \\

    id & 100.0 \scriptsize{(0.0)} & 100.0 \scriptsize{(0.0)} & 66.2 \scriptsize{(-6.7)} & 67.3 \scriptsize{(-7.1)} \\

    lv & 100.0 \scriptsize{(+0.1)} & 100.0 \scriptsize{(+0.1)} & 84.2 \scriptsize{(-2.7)} & 84.1 \scriptsize{(-2.2)} \\

    mn & 100.0 \scriptsize{(0.0)} & 100.0 \scriptsize{(0.0)} & 99.7 \scriptsize{(+0.1)} & 99.9 \scriptsize{(0.0)} \\

    sl & 91.0 \scriptsize{(-5.0)} & 96.2 \scriptsize{(+0.2)} & 85.2 \scriptsize{(-3.6)} & 86.2 \scriptsize{(-3.1)} \\

    sv & 90.3 \scriptsize{(-0.6)} & 89.9 \scriptsize{(-1.1)} & 86.2 \scriptsize{(-2.6)} & 86.5 \scriptsize{(-2.6)} \\

    ta & 99.9 \scriptsize{(0.0)} & 99.9 \scriptsize{(0.0)} & 98.8 \scriptsize{(+1.7)} & 99.9 \scriptsize{(0.0)} \\

    tr & 99.3 \scriptsize{(-0.1)} & 99.5 \scriptsize{(+0.1)} & 81.9 \scriptsize{(-4.1)} & 83.5 \scriptsize{(-2.5)} \\
    
    \noalign{\global\arrayrulewidth=0.8pt}
    \hline
    \noalign{\global\arrayrulewidth=0.4pt}
    
    \end{tabular}
    \caption{Replacement rates (\%) of code-switching translations. Deltas from monolingual replacement rates are in parentheses.}
    \label{tab:replacementrate}
\end{table}

\noindent \textbf{Replacement Rates.}
As in copy rates, replacement rates are also generally lower for csw$\rightarrow$X translation than csw$\rightarrow$En translation, shown in Table \ref{tab:replacementrate}. Here, however, models demonstrate very high performance in csw$\rightarrow$X for languages such as Arabic, Persian, Mongolian, and Tamil, comparable to csw$\rightarrow$En translation. In contrast, they show worse performance in csw$\rightarrow$X with Latin scripts such as in Estonian or Turkish. We conjecture that scripts may be related to replacement rates, but leave this to be validated by future works.

Deltas from monolingual replacement rates are also reported in Table \ref{tab:replacementrate}. Replacement rates in code-switching translations are generally lower than those in monolingual translations. In the very occasional cases where code-switching translation replacement rates are higher, margins are very small, with the largest at 1.7 percent.

\begin{figure}
    \centering
    \includegraphics[scale=0.24]{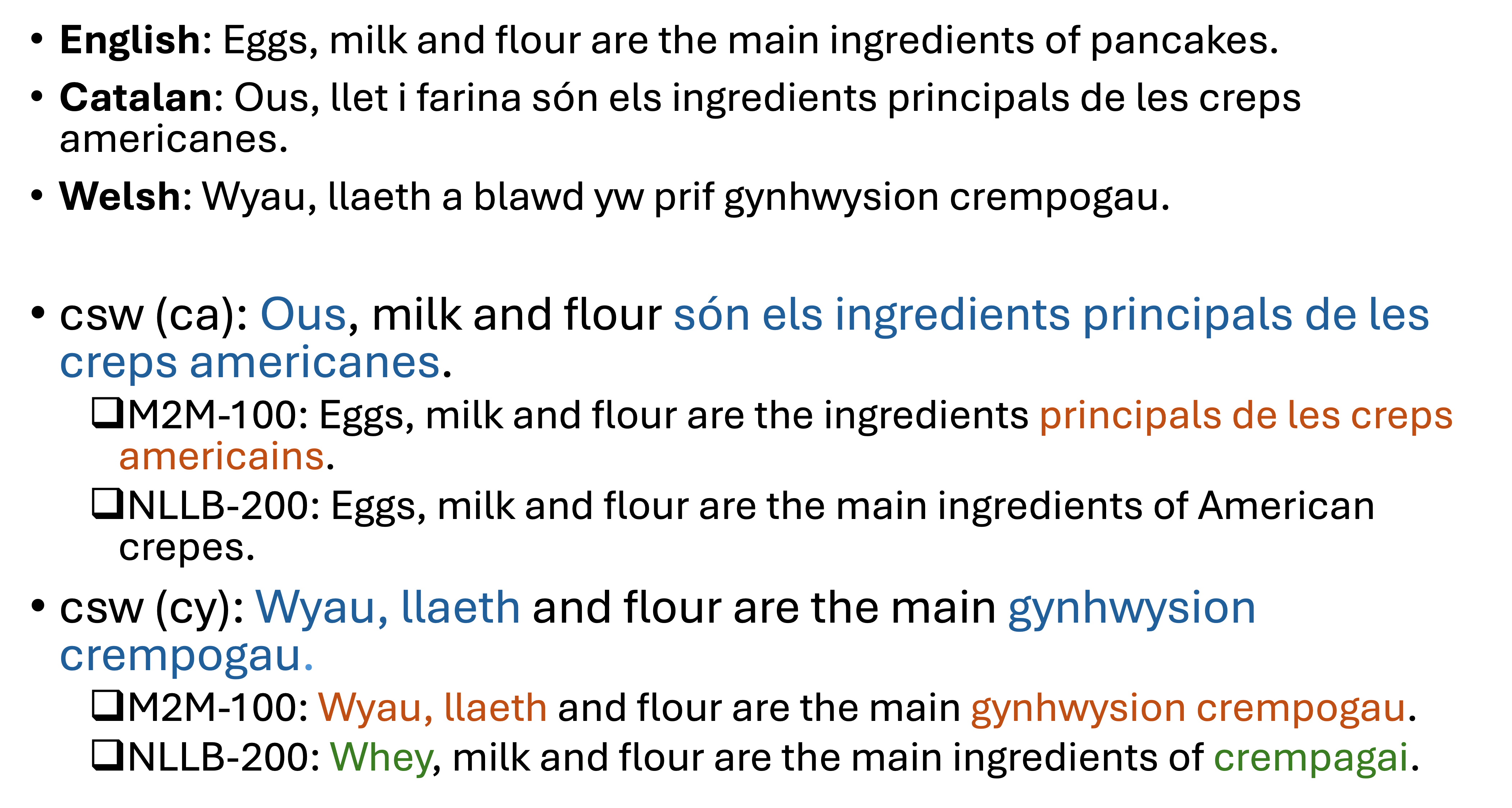}
  \caption{Example translation output in Catalan-English and Welsh-English for csw$\rightarrow$En task.}
  \label{fig:CSTranslation}
\end{figure}

\noindent \textbf{Off-target Problem and Hallucination.} Low replacement rates in csw$\rightarrow$X translation suggest that a considerable fraction of words are not being translated, despite target language being specified. Table \ref{tab:replacementrate} indicates that up to 33.8\% of English tokens are not translated into Indonesian with M2M-100 and up to 32.7\% of English tokens are not translated into Indonesian with NLLB-200. Figure \ref{fig:CSTranslation} shows examples of fully and partially translated system outputs in Catalan-English and Welsh-English. Words in orange are code-switched tokens that remain in the system output of multilingual machine translation models. We believe this points to a case of the off-target problem seen in massively multilingual translation models \citep{Zhang:20, Liu:23, Chen:23, Guerreiro:23}, studied primarily in monolingual translation settings thus far. In our code-switching translation experiments, models ignore the specified target language and instead copy the code-switched input as the translation output.

\begin{figure}
    \centering
    \includegraphics[width=0.4\textwidth]{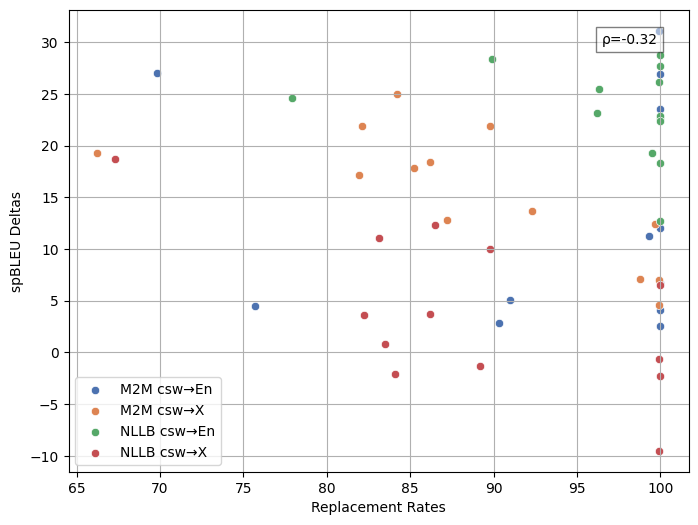}
\caption{Replacement rates plotted against spBLEU deltas. Correlation $\rho$ in the upper right corner is measured with Spearman's coefficient.}
\label{fig:correlation}
\end{figure}

Recent work \citep{Tan:23} demonstrated that the off-target problem is a symptom rather than a cause of poor zero-shot translation in monolingual settings. To understand this in the code-switching context, we apply their methods and measure the correlation between replacement rates and spBLEU deltas relative to raw code-switched inputs, shown in Figure \ref{fig:correlation}. While there is a slight negative correlation, spBLEU deltas for replacement rates of 100\% vary significantly. We therefore conclude that replacement rates are likewise not direct causes of poor code-switching translation, in accordance with prior findings.

Figure \ref{fig:CSTranslation} also illustrates a case of hallucination. In the Welsh-English NLLB-200 translation, the words in green, \textit{Whey} and \textit{crempagai}, are absent in the original Welsh and English sentences. We observe, however, that the model attempted to translate or scramble the Welsh words given the similarity of \textit{Wyau} and \textit{Whey} and \textit{crempogau} and \textit{crempagai}. In addition, this demonstrates the off-target problem as models were tasked with translation into English. Hallucinations observed in csw$\rightarrow$X translation are included in Appendix \ref{subsec:off-target-csw-x}.

\section{Conclusion}

In this work, we present CoVoSwitch, a code-switching dataset created by replacing intonation units detected by PSST, a speech segmentation model fine-tuned from Whisper, on CoVoST 2, a speech-to-text translation dataset. Using CoVoSwitch, we examine the performance of two MNMT models with millions of parameters, M2M-100 418M and NLLB-200 600M, and compare code-switching translations against monolingual translations and high-resource languages against low-resource languages. We discover that the introduction of code-switching units results in higher performing translations compared to monolingual settings and that models are better at code-switching translation into English than into non-English. Meanwhile, low-resource languages gain most from monolingual baselines compared to other languages in csw$\rightarrow$En but much less in csw$\rightarrow$X. Systems also exhibit poor translation abilities in low-resource csw$\rightarrow$X translation to the extent that performance already gained from code-switched inputs is lost. Additionally, we find that models struggle to copy non-English tokens, identify the off-target problem in code-switching settings, and confirm that models hallucinate in code-switching translation by creating words nonexistent in the original source sentences. By releasing CoVoSwitch, we aim to support the inclusion of a wider variety of languages in code-switching research.

\section*{Limitations}
We used English as the matrix language following the Matrix Language Frame Model and detected English intonation units. Future work could explore code-switching based on intonation unit replacement on languages other than English and analyze any translation performance differences from this work. Alternative methods for intonation unit replacement could also be studied for scriptio continua languages that we excluded for cross-lingual comparative analysis.

\section*{Ethics Statement}
This work does not pose ethical issues. All datasets and models used in this study are publicly available and were used under their respective Creative Commons licenses.

\section*{Acknowledgements}
We thank the anonymous reviewers for their insightful comments and suggestions.

\bibliography{custom}

\newpage
\appendix
\section{Appendix}

\subsection{Languages in the Synthesized Dataset}
We report the ISO 639-1 code, language name, family, subgrouping, script, and resource level for the 13 languages that we incorporated from CoVoST 2 in Table \ref{tab:languages}. We draw the information on language family, subgrouping, script, and resource level from \citep{NLLBTeam:22}. \citep{NLLBTeam:22} indicates resource level with either high or low.

\label{subsec:languages}

\subsection{Statistics on Train and Validation Subsets}
\label{subsec:trainvalidation}

We include statistics on train and validation subsets of CoVoSwitch, created from the train and validation subsets of CoVoST 2 in Tables \ref{tab:synthetictrain} and \ref{tab:syntheticvalidation}.

\renewcommand{\arraystretch}{1.2} 
\begin{table}[h]
    \centering
    \small
    \begin{tabular}{cccccc}
    \noalign{\global\arrayrulewidth=0.8pt}
    \hline
    \noalign{\global\arrayrulewidth=0.4pt}
         \textbf{ISO} & \textbf{Count} & \textbf{\%L1} & \textbf{\%L2} & \textbf{CMI} & \textbf{SPF} \\
        \hline
        ar & 145,115 & 54.55 & 45.45 & 32.74 & 0.17 \\
        ca & 143,880 & 50.33 & 49.67 & 33.31 & 0.15 \\
        cy & 143,473 & 51.89 & 48.11 & 33.21 & 0.16 \\
        de & 143,851 & 50.50 & 49.50 & 33.29 & 0.15 \\
        et & 144,239 & 55.38 & 44.62 & 32.65 & 0.17 \\
        fa & 145,605 & 51.37 & 48.63 & 33.23 & 0.15 \\
        id & 143,277 & 52.68 & 47.32 & 33.19 & 0.16 \\
        lv & 145,320 & 54.32 & 45.68 & 32.81 & 0.17 \\
        mn & 145,154 & 54.50 & 45.50 & 32.78 & 0.17 \\
        sl & 144,361 & 53.35 & 46.65 & 33.09 & 0.16 \\
        sv & 143,235 & 51.93 & 48.07 & 33.10 & 0.16 \\
        ta & 145,227 & 54.73 & 45.27 & 32.83 & 0.17 \\
        tr & 144,543 & 54.82 & 45.18 & 32.88 & 0.17 \\
    \noalign{\global\arrayrulewidth=0.8pt}
    \hline
    \noalign{\global\arrayrulewidth=0.4pt}
    \end{tabular}
    \caption{Train subset of CoVoSwitch. L1 is English, L2 is non-English language indicated by the ISO code.}
    \label{tab:synthetictrain}
\end{table}

\renewcommand{\arraystretch}{1.2} 

\begin{table}[h]
    \centering
    \small
    \begin{tabular}{cccccc}
    \noalign{\global\arrayrulewidth=0.8pt}
    \hline
    \noalign{\global\arrayrulewidth=0.4pt}
         \textbf{ISO} & \textbf{Count} & \textbf{\%L1} & \textbf{\%L2} & \textbf{CMI} & \textbf{SPF} \\
        \hline
        ar & 6,784 & 54.53 & 45.47 & 32.42 & 0.17 \\
        ca & 6,717 & 50.12 & 49.88 & 32.97 & 0.16 \\
        cy & 6,684 & 51.62 & 48.38 & 32.99 & 0.16 \\
        de & 6,711 & 50.30 & 49.70 & 33.01 & 0.16 \\
        et & 6,735 & 55.07 & 44.93 & 32.40 & 0.18 \\
        fa & 6,786 & 51.43 & 48.57 & 32.91 & 0.16 \\
        id & 6,659 & 52.48 & 47.52 & 32.96 & 0.17 \\
        lv & 6,774 & 54.14 & 45.86 & 32.52 & 0.17 \\
        mn & 6,772 & 54.17 & 45.83 & 32.51 & 0.17 \\
        sl & 6,737 & 53.02 & 46.98 & 32.75 & 0.17 \\
        sv & 6,670 & 52.16 & 47.84 & 32.85 & 0.16 \\
        ta & 6,790 & 54.60 & 45.40 & 32.53 & 0.17 \\
        tr & 6,739 & 54.45 & 45.55 & 32.61 & 0.17 \\
    \noalign{\global\arrayrulewidth=0.8pt}
    \hline
    \noalign{\global\arrayrulewidth=0.4pt}
    \end{tabular}
    \caption{Validation subset of CoVoSwitch. L1 is English, L2 is non-English language indicated by the ISO code.}
    \label{tab:syntheticvalidation}
\end{table}

\subsection{Hallucinations in csw$\rightarrow$X Translation}
\label{subsec:off-target-csw-x}
\begin{figure}[h]
    \centering
    \includegraphics[scale=0.25]{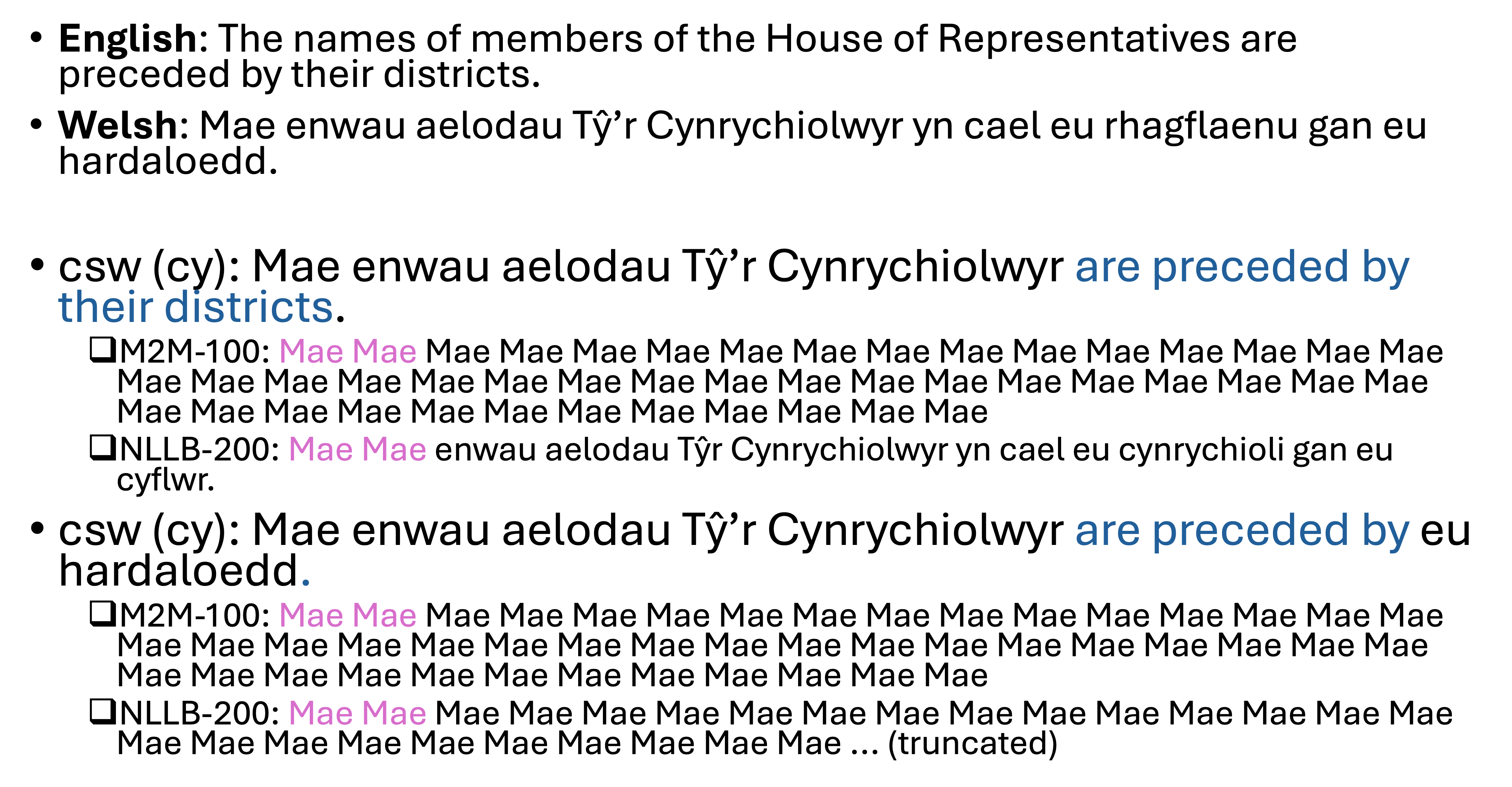}
\caption{Repeated words in csw$\rightarrow$X.}
\label{fig:csw-off-target}
\end{figure}

\begin{figure}[h]
    \centering
    \includegraphics[scale=0.25]{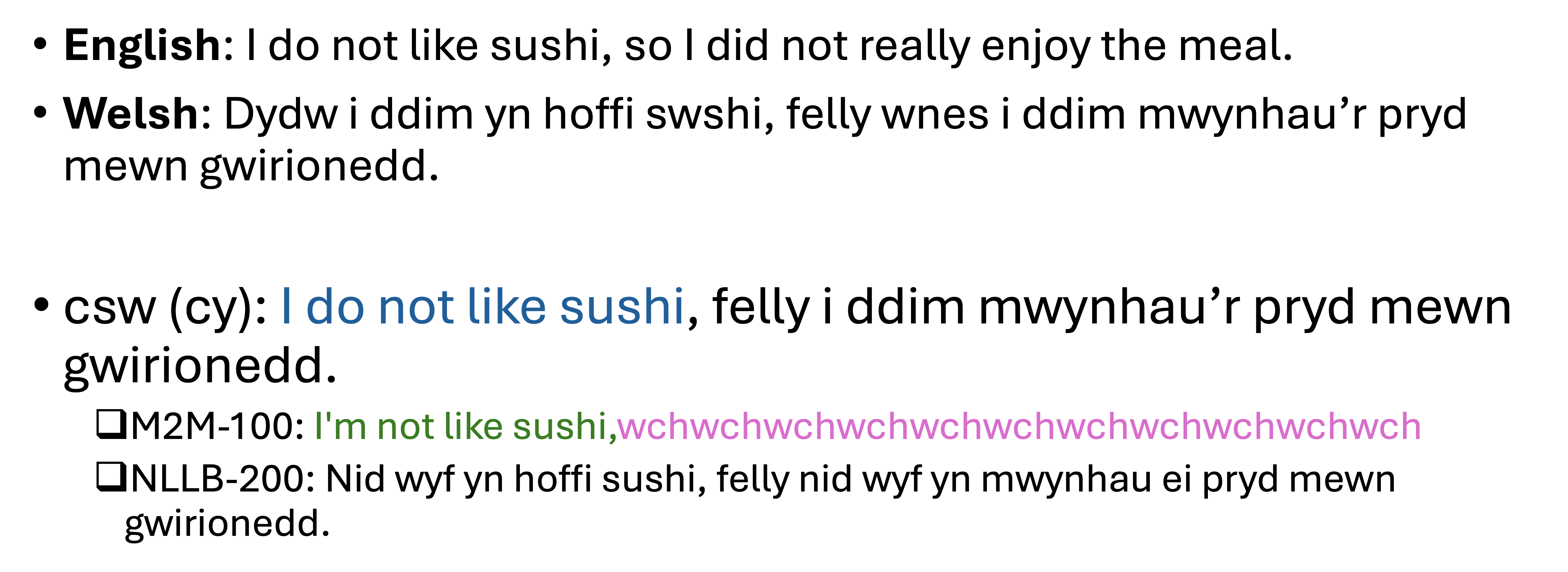}
\caption{Off-target problem, changed meaning, and repeated combinations of characters in csw$\rightarrow$X.}
\label{fig:csw-off-target-2}
\end{figure}

Hallucinations, as shown in the csw$\rightarrow$En setting in Figure \ref{fig:CSTranslation}, are also seen in csw$\rightarrow$X. As such, we provide a few observations of the problem in Welsh-English in Figures \ref{fig:csw-off-target} and \ref{fig:csw-off-target-2}. Besides the hallucination of creating words noted in Figure \ref{fig:CSTranslation}, we find repetitions of the same word. Additionally, we observe that even if two different code-switching sentences share the same source sentences, translation results can be significantly different, as seen in NLLB-200 outputs with one yielding repeated words with no meaning and the other translated but also including the repeated word \textit{Mae}, highlighted in pink.

Besides repetition of words, single characters or specific combinations of characters can be repeated, as highlighted in pink in Figure \ref{fig:csw-off-target-2}. We note that the combination repeated here, \textit{wch}, is absent in both English and Welsh source sentences and does not hold meaning relevant to the context. We find that M2M-100 not only fails to translate the English portion of the text but also completely changes its meaning when translating, from \textit{I do not like sushi} to \textit{I'm not like sushi}. This is also an example of the off-target problem because of the failure of the model to translate English to Welsh.

\subsection{Parallel Examples of Code-Switching Sentences Generated}

\begin{table*}[t]
    \small
    \centering
    \begin{tabular}{cccccc}
    \noalign{\global\arrayrulewidth=0.8pt}
    \hline
    \noalign{\global\arrayrulewidth=0.4pt}
    \textbf{ISO} & \textbf{Language} & \textbf{Family} & \textbf{Subgrouping} & \textbf{Script} & \textbf{Resource} \\
    \hline
    ar & Arabic & Afro-Asiatic & Semitic & Arabic & High \\
    ca & Catalan & Indo-European & Italic & Latin & High \\
    cy & Welsh & Indo-European & Celtic & Latin & Low \\
    de & German & Indo-European & Germanic & Latin & High\\
    et & Estonian & Uralic & Finnic & Latin & High \\
    fa & Persian & Indo-European & Iranian & Arabic & High \\
    id & Indonesian & Austronesian & Malayo-Polynesian & Latin & High \\
    lv & Latvian & Indo-European & Balto-Slavic & Latin & High \\
    mn & Mongolian & Mongolic-Khitan & Mongolic & Cyrillic & Low \\
    sl & Slovenian & Indo-European & Balto-Slavic & Latin & High \\
    sv & Swedish & Indo-European & Germanic & Latin & High \\
    ta & Tamil & Dravidian & South Dravidian & Tamil & Low \\
    tr & Turkish & Turkic & Common Turkic & Latin & High \\
    \noalign{\global\arrayrulewidth=0.8pt}
    \hline
    \noalign{\global\arrayrulewidth=0.4pt}
    \end{tabular}
    \caption{Languages used in this study in alphabetical order of ISO Code. Information on language family, subgrouping, script, and resource level is drawn from \citep{NLLBTeam:22}.}
    \label{tab:languages}
\end{table*}

\begin{figure*}[t]
    \centering
    \includegraphics[scale=0.4]{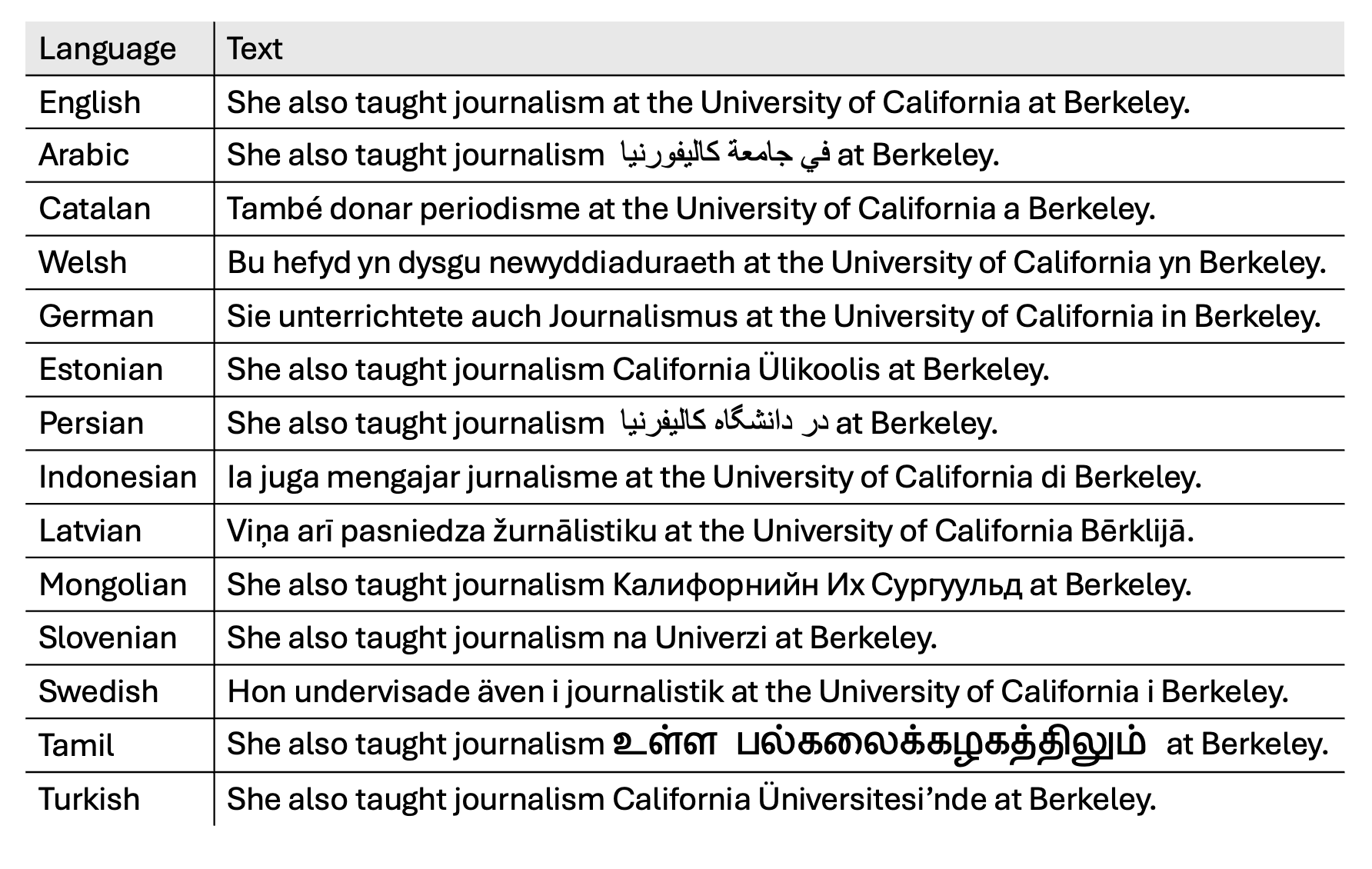}
\caption{Example of parallel code-switched text in CoVoSwitch.}
\label{fig:parallel}
\end{figure*}

All code-switched texts in CoVoSwitch are made from parallel corpora in the En$\rightarrow$X subset of CoVoST 2, and so are created using the same set of English sentences. As a result, code-switched sentences across languages share English fragments. We include an example from the test subset in Figure \ref{fig:parallel}. For some languages, we demonstrate different intonation unit replacements than others to illustrate how resulting code-switched texts diverge based on which intonation units are selected.

\end{document}